\documentclass[runningheads]{llncs}
\usepackage[T1]{fontenc}
\usepackage{graphicx}
\usepackage{makecell}

\usepackage{pifont} 
\usepackage{soul}
\usepackage{url}
\usepackage[hidelinks]{hyperref}
\usepackage[utf8]{inputenc}
\usepackage[small]{caption}
\usepackage{graphicx}
\usepackage{amsmath}
\usepackage[dvipsnames]{xcolor}
\usepackage{booktabs}
\usepackage{algorithm}
\usepackage{algorithmic}
\usepackage{float}
\usepackage[table]{xcolor}
\usepackage{amssymb}
\usepackage{array}
\usepackage{multicol}
\usepackage{tikz}
\usetikzlibrary{arrows.meta, positioning, shapes.multipart}

\usepackage[table]{xcolor}
\definecolor{ok}{HTML}{DDEFE3}        
\definecolor{maybe}{HTML}{F6E7C8}     
\definecolor{no}{HTML}{F4D6D2}        
\definecolor{stripe}{HTML}{F7F7F7}    

\usepackage[table]{xcolor}
\definecolor{ok}{HTML}{D9EEF2}        
\definecolor{maybe}{HTML}{F3E6C6}     
\definecolor{no}{HTML}{F4D6D2} 
\definecolor{stripe}{HTML}{F8F8F8}

\newcommand{\cmark}{\ding{51}} 
\newcommand{\xmark}{\ding{55}} 

\usepackage{latexsym}

\begin{document}
\title{CLARK: Closed-loop Learning for Adaptive Reasoning over Knowledge Graphs}

%
\titlerunning{A Probabilistic Framework for Knowledge-Driven Classification}

%

\author{Yousef Khan,\inst{1,2} \and
Luca Gherardini\inst{1}\and
Marco Maratea\inst{3}\and
Joel Arrais,\inst{2}\and
Jose Sousa,\inst{1}}
\authorrunning{Y. Khan et al.}
%
\institute{Computational Intelligence Team, Sano - Centre for Computational Personalised Medicine, Krakow, Poland  \and
CISUC/LASI, Department of Informatics Engineering, University of Coimbra, Coimbra, Portugal\and
DeMaCS, University of Calabria, Calabria, Italy
}

%

%
\maketitle              
\begin{abstract}

Machine Learning models are widely used for automating classification tasks by extracting statistical patterns from data.
However, their performance deteriorates if the data distribution changes, making them ill-suited to handle uncertain and evolving information. 
Moreover, they provide limited support for integrating prior knowledge. 
To address these limitations, we present CLARK (Closed-loop Learning for Adaptive Reasoning over Knowledge Graphs), a framework that integrates knowledge graphs, symbolic rule mining, and probabilistic reasoning under the Logic Programs with Markov Logic Networks (LP$^{\text{MLN}}$) formalism. 
Starting from CACTUS-derived KGs, CLARK translates graph structure into an LP$^{\text{MLN}}$ program and iteratively enriches it with candidate rules proposed by symbolic learners.
These rules are calibrated through probabilistic weight learning, enabling reasoning under uncertainty and refinement of the underlying graph structure.
We evaluate CLARK on two medical datasets, analysing both rule quality and downstream classification performance. 
Results demonstrate that CLARK leads to improved classification performance and more generalisable inference. 
Overall, CLARK provides a principled approach to constructing adaptive, interpretable, knowledge-driven models for classification.

\end{abstract}

\section{Introduction}
\label{sec:introduction}

Classification is one of the main objectives of modern Machine Learning (ML) and decision-support systems. 
It consists of making algorithms learn from examples and apply the extracted patterns to unseen data. 
Standard data-driven classifiers, particularly deep neural networks, achieve strong performance when trained and tested on similar distributions; the independent and identically distributed (i.i.d.) assumption \cite{datashift,rudin2019stopexplainingblackbox,shortcut}. 
However, when the distribution changes over time (distribution shift), their performance degrades, which is one of the central motivations for continuous learning in ML~\cite{datashift}.
Furthermore, ML models may rely on ``shortcut'' correlations that do not reflect the intended task and, therefore, undermine their generalisation capabilities \cite{shortcut}; this is typical of black-box models, and they are difficult to align with explicit domain knowledge or safety constraints \cite{rudin2019stopexplainingblackbox}. 
As a result, their behaviour is often bound to statistical regularities present in the training data rather than constrained by articulated knowledge about how the world works \cite{geirhos2020shortcut,lapuschkin2019unmasking}. 
This creates a gap between purely statistical classifiers and human decision-making, which routinely exploits background knowledge, empirical evidence, and counterfactual reasoning~\cite{ye2025cleverhansmiragecomprehensive,Zhang_2019}. 
To narrow this gap, we seek architectures that treat classification not only as a loss function minimisation, but as a system that is also able to reason with uncertainty and structured knowledge.

Knowledge graphs (KGs) provide a natural representation of structured knowledge that can be incorporated with reasoning under uncertainty in classification tasks. 
They encode taxonomies, causal relations, domain constraints, and multi-relational structure in a machine- and human-readable format \cite{damato_et_al:TGDK.1.1.8,Hogan2021KGs}. 
When used in classification, KGs can be injected as domain constraints (e.g., mutual exclusivity, temporal ordering), enrich feature spaces with relational context, and support explanation by tracing predictions back to graph paths or rules \cite{SCHRAMM2023100806}. 
They can also reduce data requirements by enabling models to generalise through a shared relational structure, allowing knowledge acquired from one part of the graph to inform reasoning in others, rather than relying solely on large quantities of labelled training data \cite{getoor2007introduction,nickel2015review,richardson2006markov}.
This mirrors human decision-making, where observations are interpreted against a rich background model of the world, and where reasoning over entities and their relations drives classification.

To turn this structured knowledge into actionable and uncertainty-aware reasoning, we adopt a framework combining Logic Programs (LPs) with Markov Logic Networks (MLNs) called LP$^{\text{MLN}}$, a probabilistic extension of Answer Set Programming (ASP) that attaches log-linear weights to rules under the stable model semantics \cite{lee2015markov,lee2016weighted}. 
LP$^{\text{MLN}}$ combines the expressivity of non-monotonic LPs, characterised by defaults, exceptions, and constraints, with the quantitative semantics of MLNs \cite{lee2015markov,lee2016weighted,lee2018weightlearningprobabilisticextension,lee2017lpmln}. 
Soft rules encode preferences rather than hard logical constraints, allowing the system to reason under noise, incomplete information, and conflicting evidence, while still producing stable models that behave like classical answer sets in the limit of infinite weights \cite{lee2016weighted}. 
Crucially, LP$^{\text{MLN}}$ supports both inference and weight learning \cite{lee2018weightlearningprobabilisticextension}, making it well-suited to integrate prior symbolic knowledge with data-driven refinement. 
In our setting, KGs supply the structural ``world model'', while LP$^{\text{MLN}}$ provides a principled mechanism for reasoning and learning over these structures under uncertainty.

Building on this perspective, we leverage CACTUS (Comprehensive Abstraction and Classification Tool for Uncovering Structures) to extract structured knowledge from tabular data \cite{gherardini2024cactus}. 
CACTUS constructs directed KGs capturing conditional dependencies among discretised attributes, which we interpret as domain-specific representations of structured knowledge.
We interpret these KGs as domain-specific world models and translate them into LP$^{\text{MLN}}$ programs in which nodes correspond to attribute-value atoms and edges represent probabilistic rules. 
Thus, leading to our closed-loop framework, CLARK: Closed-loop Learning for Adaptive Reasoning over Knowledge graphs, which augments CACTUS-derived KGs with candidate rules mined from data and iteratively refines both rule weights and graph structure through probabilistic learning using LP$^{\text{MLN}}$. 
By combining data-driven KGs, symbolic rule learning, and probabilistic reasoning under uncertainty, CLARK provides a unified approach to knowledge-driven classification.
This approach represents a step toward adaptive, interpretable, and knowledge-driven learning systems, bridging the gap between data-driven models and symbolic reasoning in the context of KGs.

This paper makes the following contributions:
\begin{enumerate}
\item Proposes CLARK as a framework that integrates CACTUS-derived KGs, symbolic rule mining, and probabilistic reasoning under the LP$^{\text{MLN}}$ formalism.
\item Presents a principled translation from weighted KGs to LP$^{\text{MLN}}$ programs, mapping nodes and edges to soft weighted rules under stable model semantics.
\item Introduces a Structural Learner, a rule miner that exploits connected components of the CACTUS KGs, scores them using held-out log-likelihood improvement, and proposes statistically relevant Horn-style rules tailored for integration into LP$^{\text{MLN}}$.
\item Compares multiple explainable symbolic rule learners within a unified LP$^{\text{MLN}}$ learning loop, enabling rule acquisition, probabilistic calibration,  KG refinement, and downstream performance.
\end{enumerate}

A review of the relevant literature is given in Sections~\ref{sec:related-work} and \ref{sec: background}. Section~\ref{methods} describes our CLARK framework in detail. 
Section~\ref{results} empirically evaluates the framework by comparing rule learners on the quality of the rules they yield and their downstream classification performance. Section \ref{discussion} discusses the implications of refining the KG structure within LP$^{\text{MLN}}$ framework and the practical impact of CLARK.
Lastly, Section \ref{conclusions} draws conclusions and directions for future research.

\section{Related Work}
\label{sec:related-work}

This section situates our work within the broader landscape of probabilistic reasoning over KGs, symbolic rule learning, and PASP.

\paragraph{\textbf{Probabilistic reasoning over KGs.}}
A range of approaches combine logical structure with uncertainty for reasoning over KGs \cite{getoor2007introduction,kimmig2012short,richardson2006markov}. Probabilistic Soft Logic (PSL) models relational dependencies using weighted first-order rules and performs inference via convex optimisation over continuous truth values \cite{kimmig2012short}. MLNs define log-linear probability distributions over first-order formulas, enabling the integration of logic and probabilistic graphical models \cite{richardson2006markov}. While effective, these approaches either rely on continuous relaxations or lack support for non-monotonic reasoning, limiting their ability to represent defaults, exceptions, and cyclic dependencies in a declarative manner \cite{bach2017hinge,getoor2007introduction}.

\paragraph{\textbf{Rule learning over relational data and KGs.}}
A large body of work focuses on mining interpretable rules from structured data \cite{Bach,Wu2023RuleLearningKGReview}. Association rule learners such as CMAR and KG-specific systems such as AMIE3 generate Horn-style rules from observed patterns \cite{amie,cmar,Wu2023RuleLearningKGReview}. These methods are effective at discovering relational dependencies, but typically rely on heuristic metrics such as support and confidence, and are not tightly integrated with probabilistic reasoning frameworks. As a result, the learnt rules are often evaluated independently of the downstream reasoning model in which they are used.

\paragraph{\textbf{Probabilistic Answer Set Programming (PASP).}}
Existing PASP approaches commonly attach probabilities to facts or choices, inducing distributions over program instantiations and their corresponding stable models \cite{cozman2020joy,wang2015handling}. While suitable for modelling stochastic events, these approaches provide limited support for learning and calibrating dense collections of soft weighted rules under stable model semantics\cite{novelframepasp,cozman2020joy}. 


Table~\ref{plp-comparison-heatmap}  summarises key differences between major PASP frameworks~\cite{dundua2016overview,hahn2022plingo,lee2015markov,nickles2016tool}. 
The comparison highlights that LP$^{\text{MLN}}$ uniquely combines direct support for soft weighted rules, parameter learning, and compatibility under stable model semantics , while also supporting cyclic dependencies and expressive constraints. In contrast, other frameworks such as P-log and PrASP either lack native support for rule weight learning or primarily focus on probabilistic facts rather than rule-level uncertainty \cite{dundua2016overview,hahn2022plingo,lee2015markov,nickles2016tool}. This makes LP$^{\text{MLN}}$ particularly suitable for our setting, where the objective is to learn, calibrate, and refine structured dependencies represented as rules, rather than simply modelling stochastic facts.

\begin{table}[ht]
\caption{Comparison of PASP frameworks across key features:
    support for soft weighted rules, cyclic dependencies, weight learning, stable model semantics, expressive constraints, and scalable inference. \cmark = Built-in support; \xmark = Not natural/not optimised; \xmark \xmark = No support/inconsistent behaviour.}
    \centering
    \setlength{\tabcolsep}{5pt}
    \renewcommand{\arraystretch}{1.2}

    \begin{tabular}{lcccc}
        \toprule
        \textbf{Feature} &
        \textbf{LP$^{\text{MLN}}$} &
        \textbf{PrASP} &
        \textbf{P-log} &
        \textbf{plingo} \\
        \midrule

        \rowcolor{gray!5}
        \textbf{Soft rules} &
        \cellcolor{ok}\cmark &
        \cellcolor{ok}\cmark &
        \cellcolor{maybe}\xmark &
        \cellcolor{ok}\cmark \\

        \rowcolor{gray!5}
        \textbf{Cycles} &
        \cellcolor{ok}\cmark &
        \cellcolor{ok}\cmark &
        \cellcolor{maybe}\xmark &
        \cellcolor{ok}\cmark \\

        \rowcolor{gray!5}
        \textbf{Learning} &
        \cellcolor{ok}\cmark &
        \cellcolor{maybe}\xmark &
        \cellcolor{no}\xmark\xmark &
        \cellcolor{no}\xmark\xmark \\

        \rowcolor{gray!5}
        \textbf{Stable models} &
        \cellcolor{ok}\cmark &
        \cellcolor{ok}\cmark &
        \cellcolor{ok}\cmark &
        \cellcolor{ok}\cmark \\

        \rowcolor{gray!5}
        \textbf{Constraints} &
        \cellcolor{ok}\cmark &
        \cellcolor{ok}\cmark &
        \cellcolor{ok}\cmark &
        \cellcolor{ok}\cmark \\

        \rowcolor{gray!5}
        \textbf{Scalable} &
        \cellcolor{ok}\cmark &
        \cellcolor{maybe}\xmark &
        \cellcolor{maybe}\xmark &
        \cellcolor{ok}\cmark \\
        \bottomrule
    \end{tabular}

    \label{plp-comparison-heatmap}

\end{table}

\section{Background}
\label{sec: background}

This section introduces the theoretical foundations of our framework. 

\subsection{Logic Programs and Stable Model Semantics}
\label{sec:logicPrograms}

Let $\mathcal{A}$ be a finite set of propositional atoms. A normal logic rule is an expression of the form $a \leftarrow b_1, \ldots, b_m,$ $ \ \mathit{not}\ c_1, \ldots, \mathit{not}\ c_n,$ where $a,b_i,c_j \in \mathcal{A}$. A \textit{logic program} $\Pi$ is a finite set of such rules. 

In addition to normal rules, we consider extended ASP constructs used in practical modelling. A choice rule has the form $\{a_1, \ldots, a_k\} \leftarrow b_1, \ldots, b_m$, allowing any subset of atoms $a_1, \ldots, a_k$ to be selected when the body holds \cite{Baral_2003,gel88,lifschitz2019answer}. 
A cardinality constraint has the form $l\{a_1, \ldots, a_k\} u$, where $l$ and $u$ bound the number of atoms that can be true. These constructs are commonly used to enforce domain constraints such as mutual exclusivity and completeness \cite{Baral_2003,gel88,lifschitz2019answer}.
In our framework, cardinality constraints are used to enforce that each attribute takes exactly one state in every stable model.

Under the stable model semantics, each program induces a set of \textit{stable models}:  $SM(\Pi)$, corresponding to self-consistent interpretations of the rules \cite{Baral_2003,gel88,lifschitz2019answer}. These models capture non-monotonic reasoning, allowing conclusions to depend on the absence of information and to be revised when new knowledge is introduced. Answer Set Programming (ASP) provides a declarative framework for computing such models efficiently \cite{gebser2019multi,lifschitz2019answer}.

\subsection{Probabilistic Logic and LP$^{\text{MLN}}$}
\label{sec:lpmln}

To model uncertainty over logical rules, we adopt LP$^{\text{MLN}}$, which extends logic programs with real-valued weights \cite{azreasonerslpmlngithub,lee2015markov,lee2016weighted,lee2018weightlearningprobabilisticextension,lee2017lpmln}.

\begin{definition}[LP$^{\text{MLN}}$ Program]
\label{def:7}
An LP$^{\text{MLN}}$ program is a logic program with a finite set of $K$ soft weighted rules in the form $\Pi = \{\, w_i : r_i \mid i = 1,\ldots,K \,\},$
where each $r_i$ is a logic rule, including normal rules, choice rules, or rules with cardinality constraints and $w_i \in \mathbb{R}$ is its corresponding weight.

Each stable model $\omega \in SM(\Pi)$ is assigned a log-linear weight:
\[
W(\omega) = \sum_{(w_i:r_i)\in\Pi} w_i \, n_i(\omega),
\]
where $n_i(\omega)$ counts the satisfied instances of rule $r_i$, in $\omega$. 

The induced probability distribution is:
\[
\Pr(\omega) = \frac{1}{Z} \exp\!\left( W(\omega) \right),
\qquad
Z = \sum_{\omega' \in SM(\Pi)} \exp\!\left( W(\omega') \right).
\]
\end{definition}

This formulation allows rules to be treated as uncertain objects with learnable strengths, combining non-monotonic reasoning with probabilistic semantics.

\subsection{Knowledge Graphs and CACTUS}
\label{sec:cactus}

A KG is a directed, labelled graph $G = (V, E, \ell)$, where $V$ is a finite set of nodes representing entities or attribute-value pairs. $E \subseteq V \times V$ is a set of directed edges representing relations or dependencies between nodes and $\ell$ is a labelling function that assigns semantic labels and optionally confidence scores (weights) to nodes and edges \cite{Hogan2021KGs}.

In this work, we employ CACTUS to construct KGs from tabular data \cite{gherardini2025cactusreliabletoolearly,gherardini2024cactus,cactusaime}. CACTUS discretises continuous features into interpretable states and builds directed graphs capturing statistical associations between attribute-state pairs \cite{gherardini2024cactus}. This representation provides a transparent and data-efficient abstraction of the underlying dataset, enabling structured reasoning over relational dependencies.
Within CLARK, these CACTUS-derived KGs serve as domain-specific ``world models''. Nodes correspond to attribute-state atoms, and edges encode probabilistic dependencies that can be interpreted as soft logical rules. These graphs are subsequently translated into LP$^{{MLN}}$ programs, providing a bridge between data-driven structure extraction and probabilistic logical reasoning.

\section{Methods}
\label{methods}

\begin{figure*}[ht]
  \centering
  \includegraphics[width=0.99\textwidth]{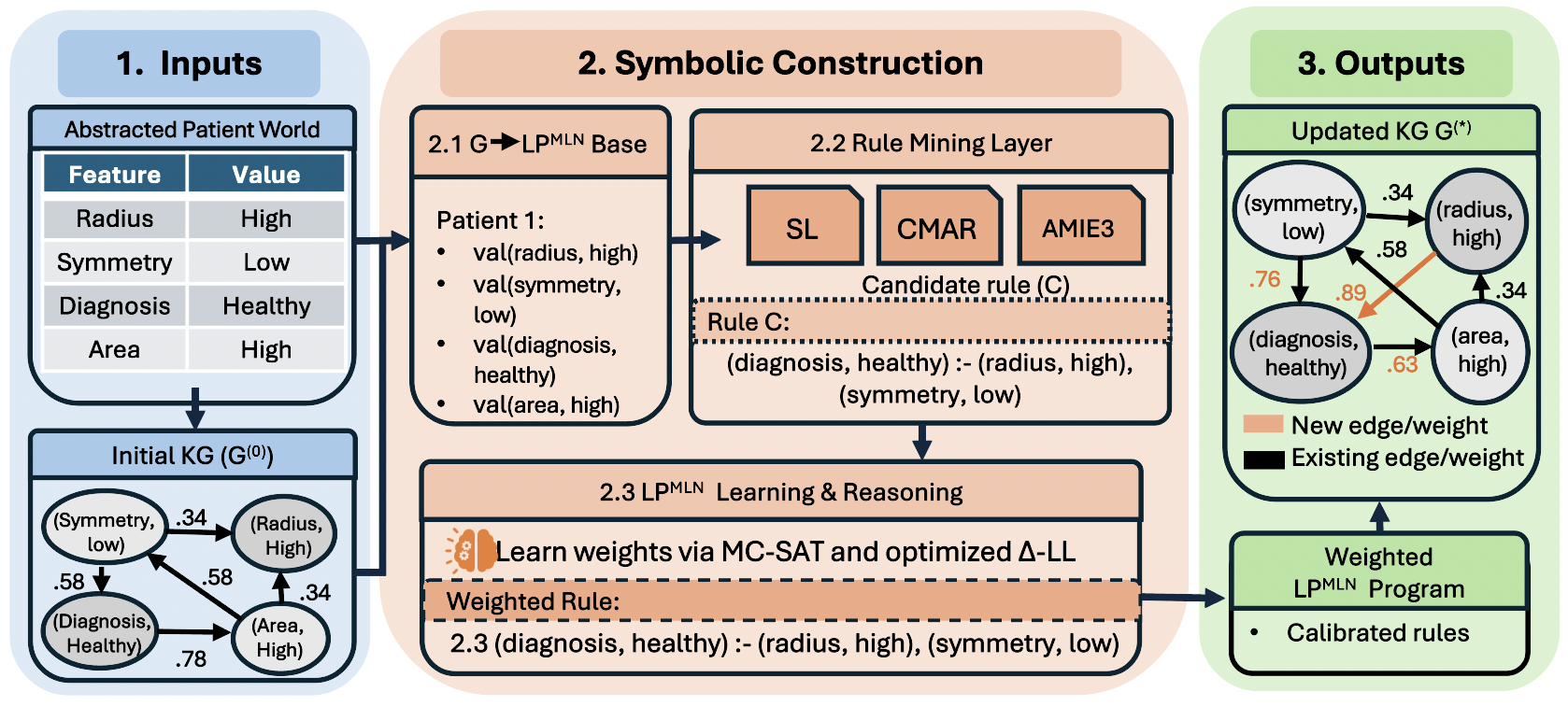} 
 \caption{\textbf{CLARK pipeline.} \textbf{(1) Inputs:} abstracted  tabular data (e.g., feature \texttt{Radius} with value \texttt{High}) and an initial KG $G^{(0)}$.
 \textbf{(2) Symbolic construction:} \textbf{(2.1)}  KG $G^{(0)}$ is translated into an LP$^{\text{MLN}}$ base program (e.g., \texttt{val(radius, high)}); \textbf{(2.2)} candidate rules are mined and \textbf{(2.3)} the base program augmented with candidate rules undergoes weight learning via MC-SAT, assigning log-linear weights. \textbf{(3) Outputs:} a weighted LP$^{\text{MLN}}$ program containing calibrated soft rules, domain constraints, inducing an updated KG $G^{(*)}$ in which accepted rules introduce new or re-calibrated weighted edges (orange) (e.g., $.78, .89$), while preserving the uncalibrated ones (black) (e.g., $.34, .58$). The learnt weight quantifies the strength of the dependency under stable model semantics and is reflected back into the enriched KG as an edge weight.}
 \label{fig:framework-overview}
\end{figure*}

We start from a CACTUS-derived KG and an abstracted tabular dataset, and aim to learn an LP$^{\text{MLN}}$ program that (i) incorporates the initial CACTUS structure, (ii) introduces additional rules proposed by explainable symbolic rule learners, and (iii) refines rule weights via probabilistic stable model learning. 
Figure~\ref{fig:framework-overview} provides a conceptual overview of the CLARK pipeline, whereas Algorithm \ref{alg:cactus-lpmln-brief} specifies the procedural steps of the iterative learning process.


\begin{algorithm}[t]
\caption{CLARK Framework}
\label{alg:cactus-lpmln-brief}
\begin{algorithmic}[1]
\REQUIRE
  Initial CACTUS graph $G^{(0)}$; discretised patient worlds $\mathcal{W}$; 
  rule learner $L \in \{\text{SL}, \text{CMAR}, \text{AMIE3}\}$; 
  maximum rounds $R$.
\ENSURE
  Enriched graph $G^{(*)}$; LP$^{\text{MLN}}$ program $\hat{\Pi}^{(*)}$.
\STATE Split $\mathcal{W}$ into $\mathcal{W}_{\text{train}}$ and $\mathcal{W}_{\text{held}}$.
\STATE $G \gets G^{(0)}$.
\FOR{$r = 1, \dots, R$}
  \STATE Build LP$^{\text{MLN}}$ base program $\Pi_{\text{base}}$ from $G$.
  \STATE $C \gets L(G,\mathcal{W}_{\text{train}},\mathcal{W}_{\text{held}})$.
  \STATE Select candidate set $\tilde{C} \subseteq C$ (e.g., top-$K$ rules).
  \STATE Form $\Pi$ by adding learnable rules \texttt{@w(k) : head :- body.} for each rule in $\tilde{C}$ to $\Pi_{\text{base}}$.
  \STATE Run LP$^{\text{MLN}}$ weight learning on $(\Pi,\mathcal{W}_{\text{train}})$ to obtain $\hat{\Pi}$.
  \STATE Apply bootstrap learning to prune unstable rules.
  \STATE Update $G$ with the learnt weights and edges induced by accepted rules.
  \IF{no edges updated}
    \STATE \textbf{break}
  \ENDIF
\ENDFOR
\STATE $G^{(*)} \gets G$, $\hat{\Pi}^{(*)} \gets \hat{\Pi}$.
\RETURN $G^{(*)}, \hat{\Pi}^{(*)}$.
\end{algorithmic}
\end{algorithm}

\subsection{From CACTUS KG to an LP$^{\text{MLN}}$ Base}
\label{sec:base-lpmln}

Starting with Part~1 in Figure~\ref{fig:framework-overview}, let $D$ be a CACTUS-abstracted tabular dataset and $G^{(0)} = (V,E)$ be the corresponding KG. Each node $v \in V$ corresponds to a discretised attribute-state pair $(a,\sigma)$ and each directed edge $(v, u) \in E$ encodes a statistical dependency. 

Part~2.1 in Figure~\ref{fig:framework-overview} introduces a principled mapping $\Phi: G \rightarrow \Pi_{base}G $, which translates a CACTUS graph $G$ into an LP$^{\text{MLN}}$ program whose stable models represent possible worlds consistent with the CACTUS KG.

\textbf{Predicate Vocabulary.}
For each attribute $a$ and state $\sigma$ observed in $G$, we introduce a propositional atom, $val(a,\sigma)$, intended to denote that attribute $a$ takes state $\sigma$ in a given world. 
Let $S(a)$ denote the set of states of attributes $a$ presents in $G$. 
Domain constraints enforce that each attribute takes exactly one admissible state in every stable model. Formally, for each attribute $a$ with state set $S(a)$, we introduce a cardinality constraint
$\texttt{1\{ val(a,$\sigma$) : $\sigma \in S(a)$ \} 1.}$
Thus, ensuring totality and mutual exclusivity of attribute assignments.

\textbf{Node Priors.}
Each node $v \in V$ may be associated with a prior score $p_v \in (0, 1)$ derived from CACTUS. 
We convert this score into a log-linear weight:  
\[
w_v = \log \frac{p_v}{1-p_v}\ .
\]
The prior information is then encoded as a soft weighted rule: 
$\texttt{$w_v : (a,\sigma)$.}$
This embeds CACTUS node importance directly into the LP$^{\text{MLN}}$ log-linear semantics while preserving stable model structure.

\textbf{Edge-Induced Rules.}
Each directed edge $(v, u) \in E$, where $v = (a, \sigma)$ and $u = (b, \tau)$, is interpreted as a soft conditional dependency. 
Let $p_{v \to u}$ denote an association strength derived from CACTUS (e.g., conditional probability or NPMI transformed into (0, 1)). 
We transform this value into a log-linear weight $w_{v\to u}$. 
The edge is translated into a weighted rule:
$w_{v\to u} \;\;\texttt{: val(a,$\sigma$) :- val(b,$\tau$).}$
Hence, the graph structure induces a set of soft Horn-style rules whose influence is calibrated in log-probability space. 

\textbf{Evidence Construction from Observations.}
Given that the dataset $D$ contains discretised patient worlds,  $\mathcal{W} = \{E^{(1)},\ldots,E^{(N)}\}$, each world $E^{(i)}$ is represented as a partial world containing observed atoms $\texttt{val}(a,\sigma)$. 
The collection of all patient worlds forms the evidence set used for training and testing (Algorithm~\ref{alg:cactus-lpmln-brief}, line~1).

\textbf{Resulting Base Program.}
The LP$^{\text{MLN}}$ base program consists of: i) Attribute domain constraints; ii) soft prior rules for nodes; iii) soft dependency rules for edges (lines ~2-4 in Algorithm \ref{alg:cactus-lpmln-brief}). 
Under  LP$^{\text{MLN}}$ semantics (Section~\ref{sec:lpmln}), this induces a probability distribution over stable models, representing complete assignments of attribute-states that respect CACTUS structural constraints while allowing uncertainty.

\subsection{Mining and Learning}
\label{sec:rule-learners}

To move beyond the pairwise structure encoded by CACTUS edges, we evaluate multiple symbolic rule learners that propose additional candidate rules over the same attribute-state atoms (Part~2.2 in Figure~\ref{fig:framework-overview}). 
All rule learners are restricted to output rules in an LP$^{\text{MLN}}$-compatible format; for instance, rules whose heads and bodies are conjunctions of \texttt{val(a,$\sigma$)} atoms. 
We consider three fully explainable symbolic rule learners: 1) A new Structural Learner (SL) we implemented for this task, 2) the Classification based on Multiple Association Rules (CMAR)~\cite{cmar}, and 3) the Association Rule Mining under Incomplete Evidence 3 (AMIE3)~\cite{amie,amiegithub}. 
These are treated as interchangeable sources for generating candidate rules. 
By keeping the LP$^{\text{MLN}}$ engine fixed and varying only the rule sources, we can statistically evaluate how different symbolic learners affect the structure of the enriched KG (Algorithm \ref{alg:cactus-lpmln-brief}, lines~5-6; Figure~\ref{fig:framework-overview}, Part 3).

\paragraph{\textbf{Structural Learner.}}
The SL introduced in this work is a rule mining method designed to perform structured learning directly under LP$^{\text{MLN}}$ stable model semantics. 
Its guiding principle is that candidate rules should be selected according to the same probabilistic objective used for weight learning by LP$^{\text{MLN}}$.

Let $Comp(G)$ denote the graphs' connected components. 
SL considers Horn-style rules of the form 
$\texttt{val(b,$\tau$) $\leftarrow$ val($a_1,\sigma_1$), ..., val($a_k,\sigma_k$),}$
subject to bounded length and that all atoms are constrained to a single connected component of $G$. 
This connectivity bias restricts the hypothesis space to graph-consistent candidates.

Given training worlds $\mathcal{W_{\text{train}}}$  and held-out world $\mathcal{W_{\text{held}}}$ each candidate rule $r$ is evaluated by extending the base LP$^{\text{MLN}}$ program $\Pi_{base}G$ and estimating  weight via maximum likelihood under  LP$^{\text{MLN}}$ semantics (Algorithm~\ref{alg:cactus-lpmln-brief} lines~5-8). 
Rule utility is measured by the held-out log-likelihood gain

$\Delta(r) =
\mathcal{L}\bigl(\Pi_{\text{base}}G \cup \{r\} \mid \mathcal{W}_{\text{held}}\bigr)
-
\mathcal{L}\bigl(\Pi_{\text{base}}G \mid \mathcal{W}_{\text{held}}\bigr).$
Only rules with positive (optionally regularised) gain are retained.


SL can thus be viewed as a constrained maximum-likelihood structural learner for probabilistic stable model programs. 
It differs from standard association rule mining in two fundamental ways. 
1) Semantic alignment: rule selection is driven by LP$^{\text{MLN}}$ likelihood rather than heuristic frequency measures. 2) Structural coherence: candidate generation respects the connected components structure induced by CACTUS. 
By unifying structural bias with probabilistic semantics, SL produces compact rule sets that integrate naturally into LP$^{\text{MLN}}$.

\subsection{LP$^{\text{MLN}}$ Reasoning and Structural Update}
\label{sec:learning-update}

Given the base program (i.e., priors and CACTUS edges) and a set of candidate rules from a learner, we obtain an LP$^{\text{MLN}}$ program $\Pi(\mathbf{w})$ with parameter vector $\mathbf{w} = (w_1,\ldots,w_K)$ indexing all $K$ candidate rules. 
Given $\mathbf{w}$, the LP$^{\text{MLN}}$ semantics define a probability distribution over the set of stable models $\omega \in SM(\Pi)$, $Pr_{\mathbf{w}} : SM(\Pi) \rightarrow [0,1]$, with $Pr_{\mathbf{w}}$ given by the LP$^{\text{MLN}}$ distribution. 
Maximum-likelihood estimation seeks weights
\[
\mathbf{w}^{\ast} \in \arg\max_{\mathbf{w}}
\sum_{i=1}^{T} \log Pr_{\mathbf{w}}(E^{(i)}),
\]
where $Pr_{\mathbf{w}}(E^{(i)})$ denotes the marginal probability of the world $E^{(i)}$, obtained by summing $Pr_{\mathbf{w}}(\omega)$ over all stable models $\omega$ consistent with $E^{(i)}$.
Given $(\Pi,\mathcal{W}_{\text{train}})$, the LP$^{\text{MLN}}$ learner returns an instantiated program $\hat{\Pi}$ in which each \texttt{@w(k)} is replaced by a numeric weight $\hat{w}_k$ (Algorithm~\ref{alg:cactus-lpmln-brief}, line~8).

As illustrated in Part~3 of Figure~\ref{fig:framework-overview}, the learnt weights are then mapped back onto the graph structure: CACTUS node and edge weights are updated by attaching the corresponding $\hat{w}_k$ values, and candidate rules with sufficiently large, stable weights are promoted to new graph edges. 
We then perform a bootstrap-style stability selection step: The training worlds are resampled, weights are re-learnt across bootstrap replicates, and only candidate rules whose learnt weights are consistently nonzero and of the same sign across replicates are retained (Algorithm \ref{alg:cactus-lpmln-brief}, line~9). 


Learning proceeds for a finite number of rounds.
At each round $r$, the current KG, $G^{(r)}$, defines the base LP$^{\text{MLN}}$ structure and the connected components within which local candidates may be mined (Algorithm \ref{alg:cactus-lpmln-brief}, lines~3-4). 
The selected rule learner produces a candidate rule set $C^{(r)}$ (Algorithm \ref{alg:cactus-lpmln-brief}, lines~5-7); which is integrated into the LP$^{\text{MLN}}$ program for weight learning over the training worlds (lines~7-8). Bootstrap pruning is then applied to remove unstable rules (line~9), and the KG is updated to obtain  $G^{(r+1)}$ (line~10). The procedure terminates when no new edges are introduced or when a preset maximum number of rounds is reached  (Algorithm \ref{alg:cactus-lpmln-brief}, lines~11-13). The final outputs are an instantiated LP$^{\text{MLN}}$ program together with an enriched KG whose edge weights and provenance reflect the accepted learnt rules (Algorithm~\ref{alg:cactus-lpmln-brief}, lines~15-16 and illustrated in Part 3 of Figure~\ref{fig:framework-overview}).

\section{Experiments}
\label{results}


This section evaluates CLARK with respect to its central objective: to improve the quality of the underlying probabilistic world model constructed from data, rather than optimising rule metrics in isolation. Specifically, we assess whether integrating symbolic rule learning with LP$^{\text{MLN}}$ reasoning leads to (i) more informative and generalisable rule sets, and (ii) improved downstream inference over the knowledge graph.

\subsection{Data}
\label{sec:data}


We evaluate our framework on two medical datasets. 
First on the Cleveland Heart Disease (HDC) dataset, a benchmark dataset for cardiovascular disease prediction  \cite{heartdataset}. The dataset contains 303 patient records described by 13 clinical and physiological attributes, including age, chest pain type, and maximum heart rate achieved. 
We additionally evaluate our framework on the Wisconsin Diagnostic Breast Cancer (WDBC) dataset, a widely used benchmark for binary classification in medical decision support \cite{wdbcdataset}. The dataset consists of 569 patient records described by 30 continuous features derived from digitised images of breast masses.
Using CACTUS, the datasets are transformed into a structured representation comprising nodes (attribute-state pairs) and directed edges encoding statistical dependencies. Each patient instance is interpreted as a \textit{world}, where each world represents a complete assignment of attribute-states consistent with the LP$^{\text{MLN}}$ program, enabling probabilistic reasoning under the LP$^{\text{MLN}}$ semantics.
This setting provides a practical use case for evaluating how structured knowledge extraction, rule learning, and probabilistic reasoning interact in a decision-support context.

\subsection{Configuration of Rule Learners}
\label{sec:rule_config}
In this subsection, we introduce our configurations of CMAR and AMIE3. All learners, including SL, are treated uniformly as external modules that propose candidate Horn-style rules over CACTUS-derived predicates, which are subsequently weighted and filtered by LP$^{\text{MLN}}$.

CMAR is an associative rule mining algorithm that builds Horn-style rules of the form $h \leftarrow b_1,\ldots,b_n$  \cite{cmar}. While CMAR was originally designed for class association rules, we configured it to mine general association rules among discretised attribute-value pairs, enabling learning beyond a single target class. Thus, the miner produces rules of the form $\texttt{val}(a_1,\sigma_1), \dots, \texttt{val}(a_k,\sigma_k) \to \texttt{class}(c)$.

AMIE3 is a Horn-rule mining system designed for large knowledge bases under the open-world assumption, where absent facts cannot be treated as negative evidence \cite{amie,amiegithub}. 
In this setting, AMIE3 was used as a pattern miner, while the resulting rule confidences were interpreted as empirical reliability scores over the observed $G$ rather than as open-world probabilities. Each AMIE3 rule was then translated into an LP$^{\text{MLN}}$ rule by turning its body into a conjunction of \texttt{val(a,$\sigma$)} atoms, translating its head into the corresponding conclusion atom, and equipping it with a learnable weight.

\subsection{Distribution of Rule Metrics}

Table \ref{tab:rule-metrics} summarises the number of rules produced by each learner and their associated runtime. CMAR generates the largest rule sets, followed by AMIE3, while SL produces substantially more compact models.
Due to exploring a smaller hypothesis space, SL performs with significantly lower computational cost. 
These results indicate that SL offers a more efficient trade-off between rule volume and computational cost, producing compact candidate sets.





\begin{table}[ht]
\centering
\caption{Number of rules contributed by each rule learner, total rules in the final LP$^{\text{MLN}}$ program, and rule-mining runtime for the HDC and WDBC datasets.}
\label{tab:rule-metrics}
\begin{tabular}{cccc|ccc}
 & \multicolumn{3}{c|}{\textbf{HDC}} & \multicolumn{3}{c}{\textbf{WDBC}} \\ 
\textbf{Metric} &
  \makecell{Total \\ Rules} &
  \makecell{Introduced \\ Rules} &
  \makecell{Runtime \\ (hh:mm)} &
  \makecell{Total \\ Rules} &
  \makecell{Introduced \\ Rules} &
  \makecell{Runtime \\ (hh:mm)} \\ \hline
\textbf{Original} & 1118 & 0 & 00:32 & 3480 & 0 & 04:16 \\
\textbf{CMAR} & 3740 & 2622 & 01:37 & 7657 & 4177 & 16:12 \\
\textbf{SL} & 2609 & 1491 & 00:50 & 4244 & 764 & 04:25 \\
\textbf{AMIE3} & 4709 & 3591 & 02:10 & 6858 & 3378 & 10:37 \\ \bottomrule
\end{tabular}
\end{table}



\begin{figure*}[h!tb]
  \centering
  \includegraphics[width=0.99\textwidth]{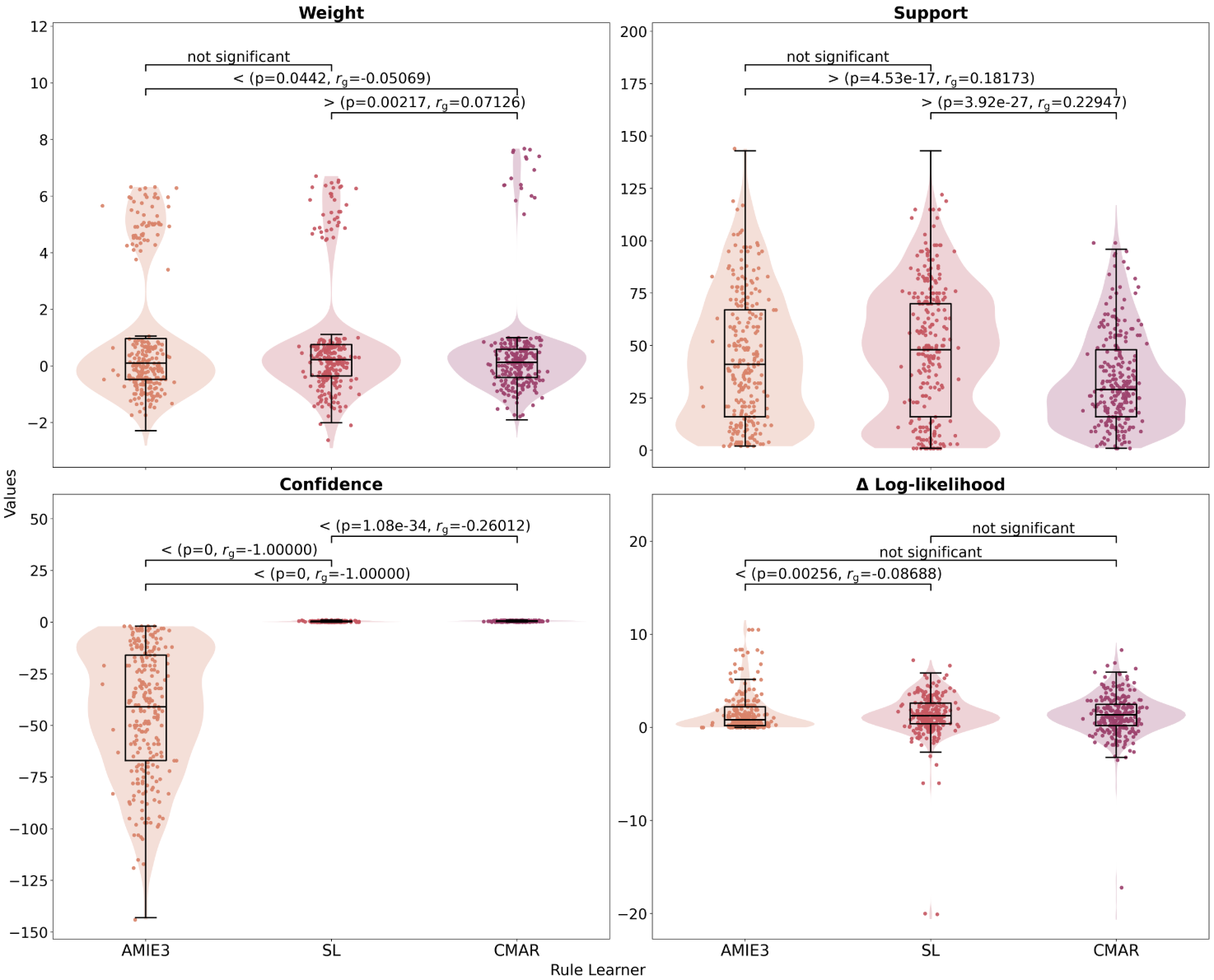}
  \caption{
Distribution of rule-quality metrics (weight, support, confidence, $\Delta$-LL) across rule learners after LPMLN learning on the HDC dataset.  
  The distributions were pairwise compared using Holm-corrected Mann-Whitney $U$ tests.}
  \label{fig:HDC-rule-metrics-fig}
\end{figure*}

\begin{figure*}[h!tb]
  \centering
  \includegraphics[width=0.99\textwidth]{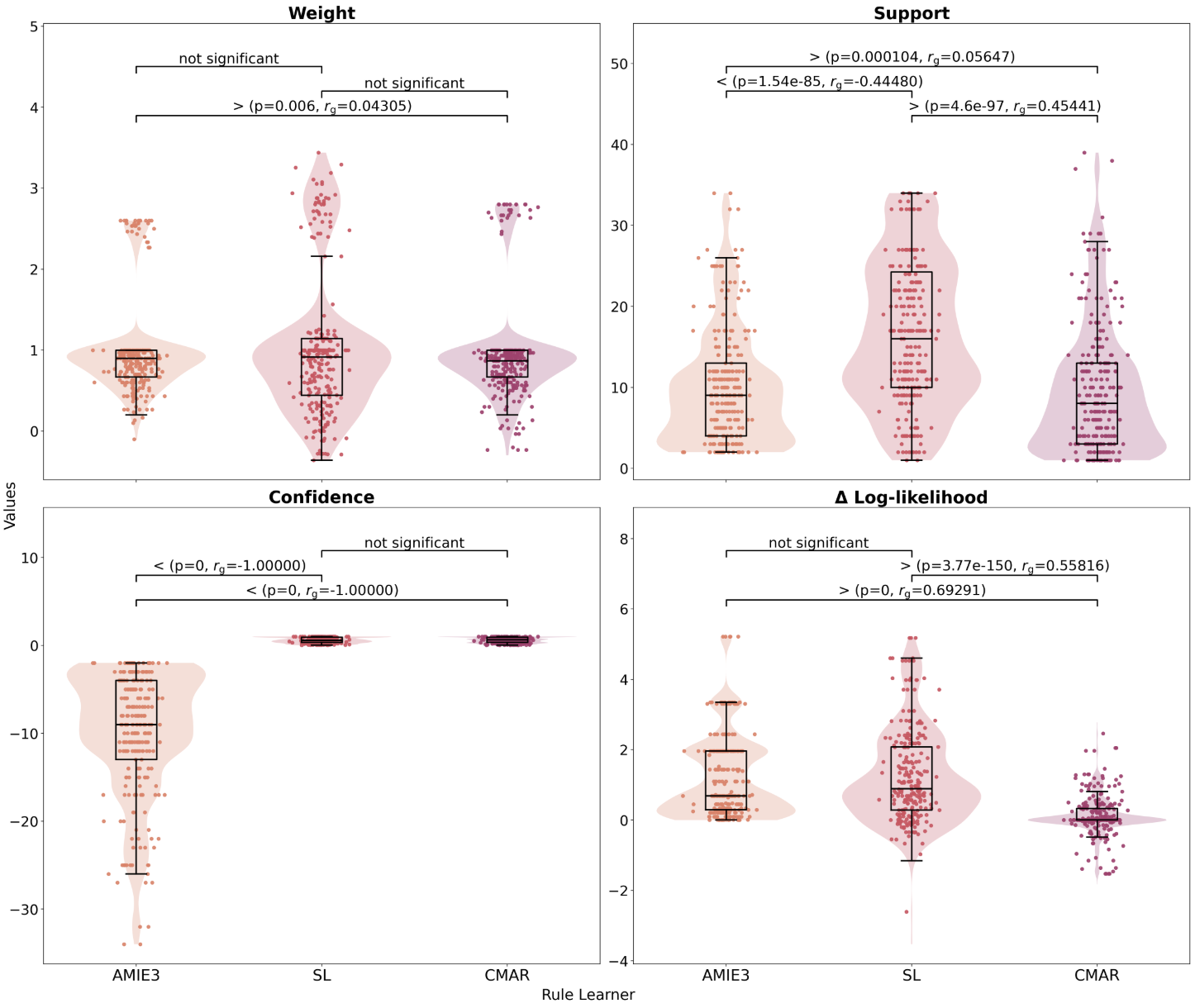}
  \caption{
Distribution of rule-quality metrics (weight, support, confidence, $\Delta$-LL) across rule learners after LPMLN learning on the WDBC dataset.  
  The distributions were pairwise compared using Holm-corrected Mann-Whitney $U$ tests.}
  \label{fig:large-rule-metrics-fig}
\end{figure*}

Figures \ref{fig:HDC-rule-metrics-fig} and \ref{fig:large-rule-metrics-fig} present violin and box plots for four rule-quality metrics (weight, support, confidence, and change in log-likelihood ($\Delta$-LL) after rule mining and LP$^{\text{MLN}}$ weight learning.  
The distributions of these metrics were tested for normality using the Shapiro-Wilk test~\cite{shapiro1965analysis}; none of them reached significance and were therefore not considered normally distributed.
Thus, differences across the mined rules were evaluated using the non-parametric Mann-Whitney U test~\cite{Mann1947}, applying Holm-Bonferroni to correct for multiple hypotheses~\cite{Holm1979}. 
This correction was preferred to the traditional Bonferroni correction to prevent excessive inflation of Type II statistical errors~\cite{Giacalone2018}.
For each comparison that reached statistical significance ($\leq 0.05$) after correction, the corresponding p-value and rank-biserial effect size are shown \cite{Willson1976}. 
The comparisons revealed significant differences across all tested rule learners.

Across both datasets, SL consistently achieves the highest median weight, support, and $\Delta$-LL (Seen in Figures \ref{fig:HDC-rule-metrics-fig} and \ref{fig:large-rule-metrics-fig}), indicating that its rules generalise across a larger fraction of patient worlds and contribute more strongly to the overall probabilistic model fit. 
In contrast, CMAR attains higher confidence scores, reflecting its bias toward frequent patterns.
Despite these differences, the distributions of learnt LP$^{\text{MLN}}$ weights are relatively similar across learners. This suggests that the probabilistic learning step effectively calibrates heterogeneous rule sources into a unified representation, normalising their influence within the final model.


\subsection{Downstream Performance}

To assess the practical impact of CLARK, we evaluate downstream classification performance for diagnosis on the HDC and WDBC datasets. In this setting, each patient corresponds to a world, and predictions are derived from the enriched KG using centrality-based scoring.

Table~\ref{tab:balanced_accuracy} reports balanced accuracy for two inference strategies: \textbf{Degree}, capturing local evidence accumulation, and \textbf{PageRank}, capturing global multi-hop influence. These results represent a comparison evaluating the contribution of each component of the CLARK framework. Starting from the original CACTUS KG as a baseline, we incrementally introduce (i) probabilistic reasoning via LP$^{\text{MLN}}$, and (ii) rule enrichment through different symbolic learners. This allows us to isolate the effect of probabilistic calibration and rule augmentation on downstream performance.
















\begin{table}[]
\caption{Classification Balanced Accuracy (BA) (\%) on the HDC and WDBC datasets. Changes are reported relative to the baseline CACTUS KG (G*) across rule-mining methods.}
\begin{tabular}{l|cc|cc|c}
 & \multicolumn{2}{c|}{\textbf{HDC}} & \multicolumn{2}{c|}{\textbf{WDBC}} & \textbf{} \\
\textbf{Classification Method} & \textbf{PageRank} & \textbf{Degree} & \textbf{PageRank} & \textbf{Degree} & \textbf{$\Delta$ BA}\\ \hline
G* & 81.16 & 80.62 & 90.97 & 92.16 & -- \\
G* + LP$^{\text{MLN}}$ & -3.83 & -4.81 & \textbf{+1.49} & -1.19 & -8.34 \\
G* + SL + LP$^{\text{MLN}}$ & +0.04 & -1.13 & +1.19 & +0.8 & +0.9 \\
G* + CMAR + LP$^{\text{MLN}}$ & +0.0 & -3.29 & +0.0 & -2.38 & -5.67 \\
G* + AMIE3 + LP$^{\text{MLN}}$ & \textbf{+1.54} & -0.97 & +0.0 & \textbf{+1.79} & +2.36 \\
G* + SL + AMIE3 + LP$^{\text{MLN}}$ & +1.22 & \textbf{+0.55} & +1.19 & +0.4 & \textbf{+3.36}\\
\end{tabular}
\label{tab:balanced_accuracy}
\end{table}



The results show that incorporating LP$^{\text{MLN}}$ reasoning alone tends to degrade predictive performance, reflecting the impact of probabilistic calibration over the original KG structure. However, these effects are not uniform across datasets: while a modest improvement is observed under PageRank for WDBC (+1.49), performance decreases in other settings, particularly for HDC. 
The introduction of learnt rules yields more consistent improvements. In particular, both SL and AMIE3 contribute to performance gains over the baseline in several settings. For WDBC, SL achieves improvements under both PageRank (+1.19) and Degree (+0.8), demonstrating stable behaviour across inference strategies. Similarly, AMIE3 provides the strongest improvement under Degree (+1.79), highlighting its effectiveness in capturing locally informative dependencies.
In contrast, CMAR does not lead to improvements and in some cases degrades performance, indicating that rules derived from frequency-based heuristics may be less aligned with the probabilistic objective of the framework.

For the combined SL+AMIE3 configuration, overlapping rules were merged by averaging their learnt metrics, while distinct rules from each learner were included without modification.
This combination of SL+AMIE3 further improves performance, achieving the highest $\Delta$ in Balanced Accuracy (+3.36).
Notably, this gain exceeds the sum of the individual  $\Delta$ in Balanced Accuracy performance from SL (+0.9) and AMIE3 (+2.36).
This suggests that agreement between the learners on the relevance of certain rules can amplify their impact, producing an effect similar to a consensus-driven update.

\section{Discussion on Experiments}
\label{discussion}

The results highlight that effective performance arises from the interaction of structural rule enrichment and probabilistic reasoning.
In particular, the combination of structured rule enrichment methods (SL + AMIE3) with likelihood-based reasoning appears essential for producing consistent gains across datasets. An important observation is that performance varies depending on the inference mechanism, indicating that different forms of reasoning (local vs. global) exploit distinct structural properties of the KG. The stability of certain configurations across both PageRank and Degree suggests that some learnt dependencies are more robust to these variations, pointing to their potential generality beyond the evaluated datasets.

More broadly, the results support the view that the quality of the underlying knowledge representation plays a central role in downstream inference. 
Rather than relying on large numbers of rules, the framework benefits from selectively incorporating dependencies that are aligned with the probabilistic objective, leading to more coherent and informative world models. 
Thus, reinforcing the central premise of the CLARK framework: that closed-loop refinement of KGs through probabilistically grounded rule learning provides a pathway toward adaptive and interpretable reasoning systems.

In a practical setting, CLARK can be used as a decision-support system in which new patient data is continuously incorporated into the KG. As new observations are added, the LP$^{\text{MLN}}$ learning process updates rule weights and introduces new dependencies, enabling the system to adapt its reasoning over time. 
For example, a newly observed correlation between tumour shape and diagnosis can be incorporated as a rule, whose weight is automatically calibrated based on its contribution to the overall probabilistic model. This allows practitioners to move beyond static predictive models toward an evolving, interpretable representation of domain knowledge that supports both prediction and explanation.

\section{Conclusions and Future Work}
\label{conclusions}

In this paper, we introduced CLARK, a closed-loop framework for learning and probabilistic reasoning over KGs derived from tabular data. The approach combines CACTUS-based KGs with probabilistic reasoning under LP$^{\text{MLN}}$, enabling structured dependencies to be represented as soft weighted rules and refined through likelihood-based learning. By integrating symbolic rule mining within this probabilistic setting, the framework supports a closed-loop process in which rule discovery, weight calibration, and graph refinement are tightly coupled.

A key strength of this framework lies in its ability to support continuous data inclusion. As new observations become available, they can be incorporated as additional worlds within the LP$^{\text{MLN}}$ learning process, triggering updates to rule weights and, when appropriate, the introduction of new dependencies in the KG. This enables the system to move beyond static representations and operate as an adaptive world model, where both the probabilistic semantics and the graph structure evolve in response to incoming data. Such capability is essential in real-world decision-support scenarios, where data distributions shift over time and models must remain robust, interpretable, and up-to-date.



Several directions for future work emerge from this study. First, we will extend the empirical evaluation to larger and more diverse datasets, including domains where relational structure plays a central role, to further assess scalability and generality. Second, we plan to investigate incremental and streaming variants of CLARK, enabling the KG and rule set to evolve dynamically as new data becomes available. Third, we aim to integrate additional rule-learning methods, including neuro-symbolic and differentiable approaches, to enrich the space of candidate rules while preserving interpretability. Finally, we will explore tighter integration with emerging PASP systems, such as plingo \cite{DBLP:journals/tplp/HahnJKRRS25}, and study how enriched KGs can support more advanced inference tasks, including explanation, counterfactual reasoning, and decision-making under uncertainty.

\paragraph{\textit{Supplemental Material Statement:}}
The source code for this framework cannot be made available due to commercialisation purposes.







\newpage

\section{Declaration of use of Generative AI}
We did not use generative AI to write or create any figures in this manuscript.

\bibliographystyle{splncs04}
\bibliography{Citations}

\end{document}